\documentclass[conference]{IEEEtran}
\usepackage{times}

\usepackage[numbers]{natbib}
\usepackage{multicol}
\usepackage{amsmath,amssymb,amsfonts}
\usepackage{algorithmic}
\usepackage{graphicx}
\usepackage{textcomp}
\usepackage{xcolor}
\usepackage{tabularx}
\usepackage{booktabs}
\newcolumntype{Y}{>{\centering\arraybackslash}X}

\usepackage{moreverb,url}
\usepackage{amsmath,amsfonts,amssymb}
\usepackage{tikz,tkz-euclide}
\usepackage{balance}
\usepackage{siunitx}
\usetikzlibrary{decorations,calligraphy}
\usetikzlibrary{calc}
\tikzset{>=latex}

\usepackage[colorlinks,bookmarksopen,bookmarksnumbered,citecolor=red,urlcolor=red]{hyperref}
\usepackage[nolist,nohyperlinks]{acronym}

\def\reals{\mathbb{R}}

\def\matern{{Mat\'{e}rn} }

\def\abscissa{\mathbf{x}}

\def\meanfun{m}
\def\point{\mathbf{p}}
\def\sensor{\mathbf{s}}
\def\range{r}
\def\max{\text{max}}

\def\coordinate{{\mathbf{e}}}

\newcommand\linedist[1]{{d^{#1}}}

\begin{acronym}[AAAAAAAAA]
    \acro{1d}[1D]{One-Dimensional}
    \acro{2d}[2D]{Two-Dimensional}
    \acro{2fast}[2Fast-2Lamaa]{Fast Field-based Agent-Subtracted Tightly-coupled Lidar Localisation And Mapping with Accelerometer and Angular-rate}
    \acro{3d}[3D]{Three-Dimensional}
    \acro{cas}[CAS]{Centre for Autonomous Systems}
    \acro{cnn}[CNN]{Convolutional Neural Network}
    \acro{cpu}[CPU]{Central Processing Unit}
    \acro{doals}[DOALS]{Urban Dynamic Objects LiDAR}
    \acro{dof}[DoF]{Degree-of-Freedom}
    \acro{dvs}[DVS]{Dynamic Vision Sensor}
    \acrodefplural{dvs}[DVS's]{Dynamic Vision Sensors}
    \acro{ekf}[EKF]{Extended Kalman filter}
    \acro{fifo}[FIFO]{First In, First Out}
    \acro{fov}[FoV]{Field-of-View}
    \acro{gnss}[GNSS]{Global Navigation Satellite System}
    \acrodefplural{gnss}[GNSS's]{Global Navigation Satellite Systems}
    \acro{gp}[GP]{Gaussian Process}
    \acrodefplural{gp}[GPs]{Gaussian Processes}
    \acro{gpm}[GPM]{Gaussian Preintegrated Measurement}
    \acro{ugpm}[UGPM]{Unified Gaussian Preintegrated Measurement}
    \acro{gps}[GPS]{Global Position System}
    \acrodefplural{gps}[GPS's]{Global Position Systems}
    \acro{gpu}[GPU]{Graphic Processing Unit}
    \acro{hdr}[HDR]{High Dynamic Range}
    \acro{hri}[HRI]{Human-Robot Interactions}
    \acro{icp}[ICP]{Iterative Closest Point}
    \acro{idol}[IDOL]{IMU-DVS Odometry using Lines}
    \acro{iou}[IoU]{Intersection over Union}
    \acro{imu}[IMU]{Inertial Measurement Unit}
    \acro{in2laama}[IN2LAAMA]{INertial Lidar Localisation Autocalibration And MApping}
    \acro{kf}[KF]{Kalman Filter}
    \acro{kl}[KL]{Kullback–Leibler}
    \acro{lidar}[LiDAR]{Light Detection And Ranging Sensor}
    \acro{lpm}[LPM]{Linear Preintegrated Measurement}
    \acro{map}[MAP]{Maximum A Posteriori}
    \acro{mle}[MLE]{Maximum Likelihood Estimation}
    \acro{ndt}[NDT]{Normal Distribution Transform}
    \acro{pca}[PCA]{Principal Component Analysis}
    \acro{pm}[PM]{Preintegrated Measurement}
    \acro{rcnn}[R-CNN]{Region-based Convolutional Neural Network}
    \acro{rrbt}[RRBT]{Rapidly Exploring Random Belief Trees}
    \acro{rgb}[RGB]{color}
    \acro{rgbd}[RGBD]{color-depth}
    \acro{rms}[RMS]{Root Mean Squared}
    \acro{rmse}[RMSE]{Root Mean Squared Error}
    \acro{ros}[ROS]{Robot Operating System}
    \acro{sde}[SDE]{Stochastic Differential Equation}
    \acro{sdf}[SDF]{Signed Distance Function}
    \acro{SE3}[SE(3)]{Special Euclidean group in three dimensions}
    \acro{slam}[SLAM]{Simultaneous Localisation And Mapping}
    \acro{so3}[$\mathfrak{so}$(3)]{Lie algebra of special orthonormal group in three dimensions}
    \acro{SO3}[SO(3)]{Special Orthonormal rotation group in three dimensions}
    \acro{upm}[UPM]{Upsampled-Preintegrated-Measurement}
    \acro{uts}[UTS]{University of Technology, Sydney}
    \acro{vi}[VI]{Visual-Inertial}
    \acro{vio}[VIO]{Visual-Inertial Odometry}
    \acro{vo}[VO]{Visual Odometry}
\end{acronym}

\begin{document}

\title{Towards Efficient Occupancy Mapping via\\Gaussian Process Latent Field Shaping}


\author{\authorblockN{Cedric Le Gentil}
\authorblockA{
University of Toronto\\
Institute for Aerospace Studies\\
Canada\\
Email: cedric.legentil@utoronto.ca}
\and
\authorblockN{Cédric Pradalier}
\authorblockA{
Georgia Tech-CNRS\\
International Research Lab 2958\\
France\\
Email: cedricp@georgiatech-metz.fr}
\and
\authorblockN{Timothy D. Barfoot}
\authorblockA{
University of Toronto\\
Institute for Aerospace Studies\\
Canada\\
Email: tim.barfoot@utoronto.ca}
}


\maketitle

\begin{abstract}
Occupancy mapping has been a key enabler of mobile robotics.
Originally based on a discrete grid representation, occupancy mapping has evolved towards continuous representations that can predict the occupancy status at any location and account for occupancy correlations between neighbouring areas.
Gaussian Process (GP) approaches treat this task as a binary classification problem using both observations of occupied and free space.
Conceptually, a GP latent field is passed through a logistic function to obtain the output class without actually manipulating the GP latent field.   
In this work, we propose to act directly on the latent function to efficiently integrate free space information as a prior based on the shape of the sensor's field-of-view.
A major difference with existing methods is the change in the classification problem, as we distinguish between free and unknown space.
The `occupied' area is the infinitesimally thin location where the class transitions from free to unknown.
We demonstrate in simulated environments that our approach is sound and leads to competitive reconstruction accuracy.
\end{abstract}

\IEEEpeerreviewmaketitle

\section{Introduction}

\begin{figure}
    \centering
    \def\imgheight{4.2cm}
    \def\hdist{0.3cm}
    \def\vdist{0.4cm}
    \def\legenddistup{-0.2cm}
    \def\legenddist{0.1cm}
    \def\legendtextsize{\small}
    \begin{tikzpicture}
        \node[inner sep=0, outer sep=0] (latent) {\includegraphics[height=\imgheight, clip, trim=1.3cm 0.3cm 1.8cm 0.9cm]{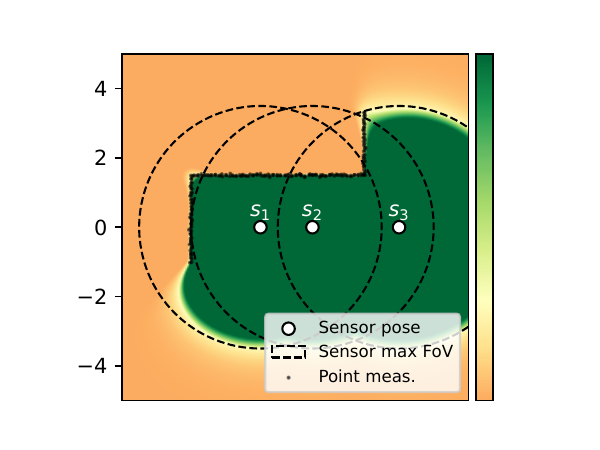}};
        \node[inner sep=0, outer sep=0, right=\hdist of latent] (occ) {\includegraphics[height=\imgheight, clip, trim=2.2cm 0.3cm 2.0cm 0.9cm]{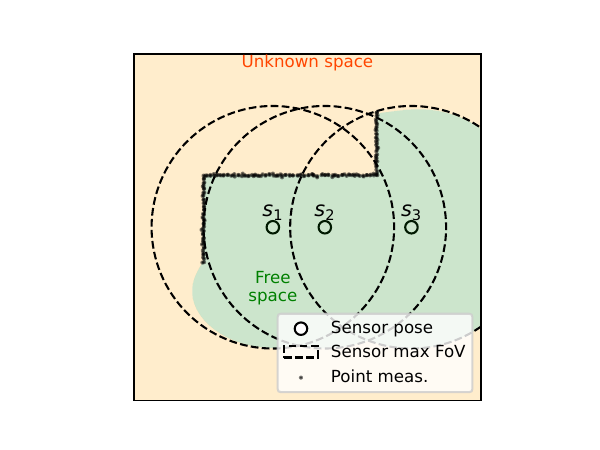}};
        \node[inner sep=0, outer sep=0, below=\vdist of latent] (var) {\includegraphics[height=\imgheight, clip, trim=1.3cm 0.3cm 1.8cm 0.9cm]{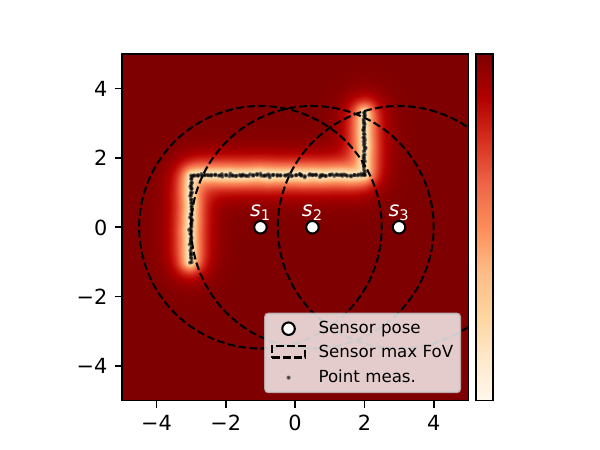}};
        \node[inner sep=0, outer sep=0, right=\hdist of var] (reconstruction) {\includegraphics[height=\imgheight, clip, trim=2.2cm 0.3cm 2.0cm 0.9cm]{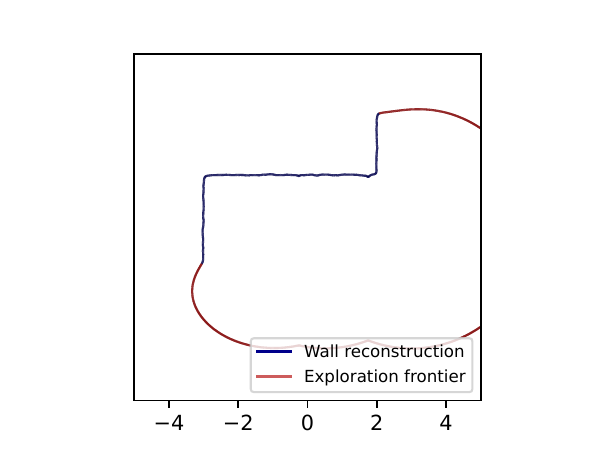}};
        \node[inner sep=0, outer sep=0, below=\legenddistup of latent]{\legendtextsize(a) GP latent field};
        \node[inner sep=0, outer sep=0, below=\legenddistup of occ]{\legendtextsize(b) Occupancy classification};
        \node[inner sep=0, outer sep=0, below=\legenddist of var]{\legendtextsize(c) Latent field variance};
        \node[inner sep=0, outer sep=0, below=\legenddist of reconstruction]{\legendtextsize(d) Surface reconstruction};
        \draw[line width=2, red] (2.1, -0.6) -- (2.3, -0.6);
        \node[rotate=-90, red] at (2.42, -0.6) {\scriptsize Level set};
    \end{tikzpicture}
    \caption{The proposed method addresses the problem of occupancy mapping as a classification problem using a continuous Gaussian Process (GP) latent field (a).
    The environment is classified between free and unknown-occupancy spaces by binarizing the latent value based on a predefined level set (b).
    The key contribution of our method is to embed knowledge of the sensor's field-of-view in the GP prior to prevent the need for explicit free-space observations.
    The variance of the latent field (c) provides information about the nature of the free/unknown transition, thus allowing the distinction between wall reconstruction and exploration frontiers (d).}
    \label{fig:teaser}
\end{figure}
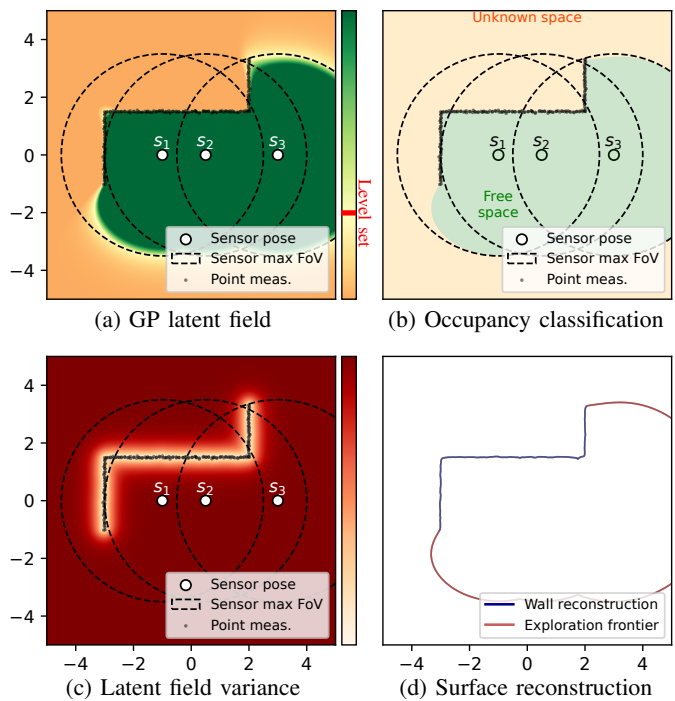

Occupancy mapping is the backbone of countless mobile robots' navigation stacks.
Since the introduction of occupancy grids~\cite{elfes1989occupancy} for 2D mapping, numerous works have built upon this concept for 3D navigation~\cite{hornung2013octomap}, including priors~\cite{merali2021mcmc}, continuous mapping~\cite{ramos2016hilbert}, etc.
While one of the main uses of occupancy mapping is the representation of terrain traversability for planning~\cite{tsardoulias2016review}, works such as~\cite{kim2012occupancy,guizilini2018towards} use it for geometric reconstruction of the environment.
In this work, we address the task of occupancy mapping using \ac{gp} regression.
Occupancy mapping can be seen as a binary classification problem.
O'Callaghan et al.~\cite{ocallaghan2009contextual} have used~\ac{gp} classification to provide a continuous representation of occupancy.
The essence of \ac{gp} classification consists in representing a \emph{latent} function $f: \reals^D \to \reals$ ($D$~the number of dimensions) with a \ac{gp} and using the logistic function to `squash' the model's output between 0 and 1 to obtain the class probability~\cite{rasmussen2006gaussian}.
Standard \ac{gp} classification does not explicitly manipulate $f$.
In this work, we propose explicitly acting on the latent function to embed prior knowledge about the sensor's characteristics.

A limitation of current \ac{gp}-based occupancy formulations is the need for both free-space and occupied-space observations.
Thus, the raw data points from range sensors alone are not sufficient to derive an occupancy function, as the measured points correspond only to occupied-space observations.
Methods such as~\cite{ghaffari2018gaussian} create virtual free-space measurements by sampling points along the range sensor's rays.
This creates additional data that increases the already-high computational burden of \ac{gp} classification (cubic complexity with respect to the number of observations).
In~\cite{OCallaghan2011}, the authors consider kernel integrals to represent the free space between the sensor and each of the individual point measurements.
While this limits the number of observations when compared to the point-sampling strategy, it still doubles the amount of data as each point is associated with an integral free-space virtual measurement.
Additionally, the derivation of the covariance kernel's integral is not trivial in many cases.

In this work, we act directly on the \ac{gp}-modelled latent function $f$ and leverage the knowledge of the sensor's \ac{fov} to represent free-space efficiently.
Our method is inspired by the implicit surface representation widely used in computer graphics to represent closed shapes with a continuous function over $\reals^{D}$.
That implicit function equals a level set~$c$ on the surface, is inferior to~$c$ inside the shape, and is superior to~$c$ outside.
Works such as~\cite{Williams2006,dragiev2011gaussian} have used \acp{gp} to model implicit surfaces.
Unfortunately, in robotics, we often have to operate with partial observations of the environment.
In~\cite{martens2017geometric}, the authors use simple geometric priors in their \ac{gp} implicit surfaces to handle partial observations of individual objects.
However, these methods do not account for the limitation of the sensor \ac{fov}, and in the context of mobile robotics, many objects of various shapes are present in the environment.
We propose to adapt the implicit surface definition by considering free-space and unknown-occupancy instead of inside and outside.
As illustrated in Fig.~\ref{fig:teaser}, our method can be seen as a \ac{gp} classification by binarizing the inferred latent function $f(\abscissa)$, at any location $\abscissa \in \reals^D$, using a pre-defined level set $c$.
The crossing of the level set corresponds to the boundary between known free-space and unknown-occupancy areas.
Analyzing the field's variance at the crossing location allows for determining whether it corresponds to a wall that has been observed by the robot or a frontier that can be used for exploration.

Standard occupancy mapping models areas as occupied or free, but the nature of ranging sensors only provides returns on the surface.
An advantage of the proposed approach with respect to existing \ac{gp}-based methods is the infinitesimal thickness of the `occupied' space.
Accordingly, our method aims to improve the surface reconstruction abilities of occupancy mapping methods while only storing a sparse parameterization of space.
Our empirical results in 2D support this claim and pave the way towards 3D reconstruction as part of our future work.

\section{Preliminaries}

As this work relies heavily on \ac{gp} regression, we provide a brief overview of the topic in this section.
For more details, we refer the reader to \cite{rasmussen2006gaussian}.
Let us consider a function $f(\abscissa): \mathbb{R}^D \rightarrow \mathbb{R}$ and noisy observations of this function at $N$ locations $y_i = f(\mathbf{x}_i) + \epsilon$, where $\epsilon \sim \mathcal{N}(0, \sigma^2)$ with $i = 1,\cdots, N$.
By modelling the function $f$ as a zero-mean \ac{gp},
\begin{equation}
    f(\abscissa) \sim \mathcal{GP}(\meanfun(\abscissa), k(\abscissa, \abscissa')),
    \label{eq:gp_model}
\end{equation}
with $\meanfun(\abscissa)$ the prior mean function and $k(\abscissa,\abscissa')$ the covariance kernel function, the posterior distribution of the function at a new location $\abscissa^*$ can be computed as
\begin{equation}
    \begin{gathered}
    p(f(\abscissa^*) | \mathbf{X}, \mathbf{Y}, \abscissa^*) \sim \mathcal{N}(f^*, {\sigma_f^*}^2), \text{ where }\\
    f^* = \meanfun(\abscissa^*) + \mathbf{k}_{\abscissa^*\mathbf{X}}[\mathbf{K}_{\mathbf{X}\mathbf{X}} + \sigma^2 \mathbf{I}]^{-1} (\mathbf{Y}-\meanfun(\mathbf{X})), \\
    {\sigma_f^*}^2 = k_{\abscissa^*\abscissa^*} - \mathbf{k}_{\abscissa^*\mathbf{X}}[\mathbf{K}_{\mathbf{X}\mathbf{X}} + \sigma^2 \mathbf{I}]^{-1} \mathbf{k}_{\mathbf{X}\abscissa^*},
    \label{eq:gp_posterior}
    \end{gathered}
\end{equation}
$\mathbf{X}$ is the matrix made up of the input locations, and $\mathbf{Y}$ is the vector of the noisy observations.
Note that for the sake of readability, we denote the kernel function evaluated between $\mathbf{X}$ and $\mathbf{X}'$ as $\mathbf{K}_{\mathbf{X}\mathbf{X}'}$ (matrix) and the kernel function evaluated between $\abscissa^*$ and $\mathbf{X}$ as $\mathbf{k}_{\abscissa^*\mathbf{X}}$ (vector).

\section{Occupancy with GP-based latent field}

\subsection{Problem statement}
\label{sec:problem_statement}

Let us define space occupancy as a classification problem between \emph{free} space and \emph{unknown-occupancy} space (later referred simply as \emph{unknown} space) by thresholding a latent function/field $f(\abscissa)$ over $\reals^D$ as
\begin{itemize}
    \item $f(\abscissa) = c$ represents the frontier between known and unknown space (often collocated with real-world objects, but not always),
    \item $f(\abscissa) > c$ represents the known free space,
    \item $f(\abscissa) < c$ represents the unknown space.
\end{itemize}
The latent field is parameterized by the data (point clouds) collected with a range sensor (e.g., from a lidar or depth camera) at known positions $\sensor_i$.
As the sensor requires a clear line of sight to measure a point $\point_{j}$, the space between the sensor position $\sensor_i$ is assumed to correspond to free space.
Concretely, the latent function should be above the level set ($f(\abscissa) > c$) when $\abscissa$ belongs to the line segment between $\sensor_i$ and $\point_{j}$.
Each point $\point_j$ is an observation of the frontier between the free and unknown spaces, thus $f(\point_{j})=c$.

\subsection{GP-based latent field 1D derivation}
\label{sec:derivation}

Our approach models the latent field $f(\abscissa)$ with a \ac{gp} as in~\eqref{eq:gp_model}.
We propose to efficiently account for the information of free space between the sensor position $\sensor_i$ and the points $\point_{j}$ by embedding an occupancy-motivated prior in the mean function $\meanfun(\abscissa)$ without the need for explicit free-space observations using the following properties:
\begin{itemize}
    \item[(i)] In the absence of point measurements, the area covered by the sensor's \ac{fov} should be classified as free ($f(\abscissa) > c$) and the rest of $\reals^D$ should be inferred as unknown space ($f(\abscissa) < c$),
    \item[(ii)] A point measurement should `pin down' the latent function to the level set, and we desire the unknown part of the latent field to behave similarly around frontiers that are induced by either an explicit measurement or the edge of the sensor's~\ac{fov}.
\end{itemize}
The key to the proposed method is selecting the correct mean and kernel functions in line with the sensor's~\ac{fov}.

We first consider the simplest measurement model: an omnidirectional sensor with a maximum range denoted $\range_{\max}$.
Thus, a point $\abscissa$ is in the~\ac{fov} if $\Vert\abscissa-\sensor_i\Vert < \range_{\max}$.
To fulfill property (ii), we need to formalize the meaning of `similar behaviour' in the unknown area.
While such a rule is ambiguous in two or more dimensions, we justify our choices of mean and kernel functions in a 1D world using a single sensor pose and point cloud.
In such a scenario with a single point measurement $s < p < s + \range_{\max}$,\footnote{The derivation can be done similarly with $s - \range_{\max} < p < s$}, (ii) implies that $\meanfun(s + \range_{\max}+x) = f^*(p + x)$ for $x>0$ (the right-hand side is defined as per~\eqref{eq:gp_posterior}).
It happens that this relationship holds if the covariance kernel is unscaled ($k(x, x') = 1$), isotropic and stationary ($k(x,x') \triangleq \kappa(\vert x-x'\vert)$), respects the product rule $\kappa(\vert x-x'\vert) = \kappa(x)\kappa(-x')$ if $x-x' > 0$, and the mean function is defined as $m(x) = \gamma k(s,x)$.
As per property (i), the value of $\gamma$ is a function of the level set: $\gamma = \frac{c}{\kappa(\range_{\max})}$.
The \matern 0.5 kernel fulfills the aforementioned requirements as demonstrated in Appendix~\ref{app:matern}.
In Fig.~\ref{fig:matern}, we illustrate property (i) with the prior mean on the left (no obstacle in the sensor's \ac{fov}).
When considering a measurement, the posterior mean is crossing the level set at the location of the measurement.
In Fig.~\ref{fig:matern}, property (ii) stipulates that the inferred prior mean for $x > 3.0$ in the left plot is a translation of the inferred posterior mean for $x>2.5$ on the right.

\begin{figure}
    \centering
    \includegraphics[width=0.99\columnwidth]{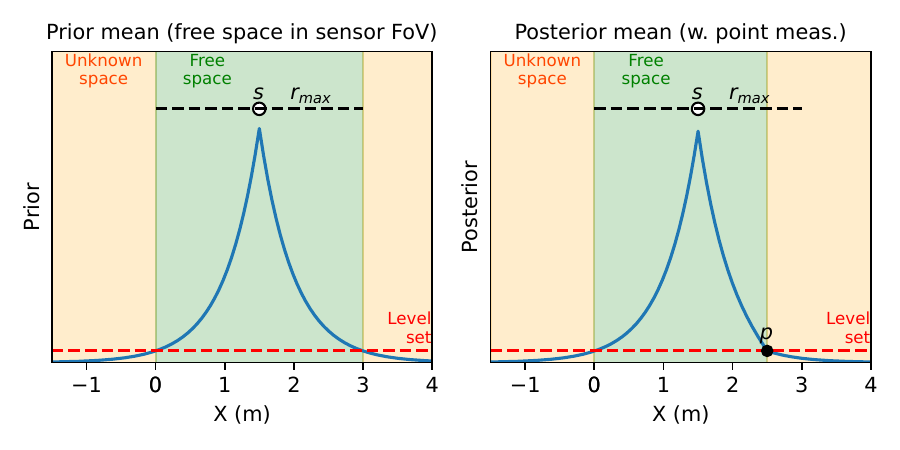}
    \caption{Illustration of the latent field properties in a 1D example with no measurement (right) and with one point measurement (right) using the \matern $\nu = 0.5$ kernel.}
    \label{fig:matern}
\end{figure}

\begin{figure}
    \centering
    \includegraphics[width=0.99\columnwidth,clip,trim=1.2cm 0.6cm 1.4cm 1.6cm]{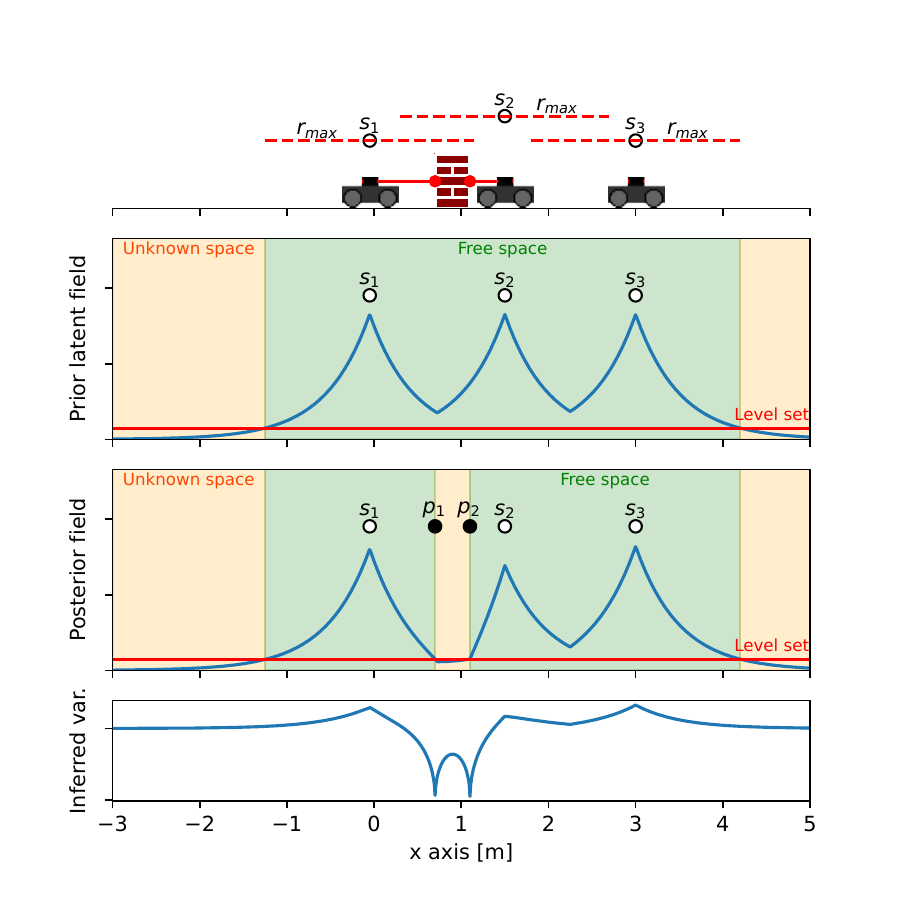}
    \caption{Considering 1D data collected by a robot at three different timestamps/poses (top row), the wall can only be sensed from $s_1$ and $s_2$ due to the limited range $\range_{\max}$ of the sensor. The proposed robot-pose-based mean function considers any area in the sensor FoV to be free space (second row). The posterior latent field (third row), using the \matern kernel with $\nu = 1/2$, accounts for the mean function and the sensor measurements $p_1$ and $p_2$. The inferred variance (last row) allows for the distinction between walls and exploration frontiers where the latent field crosses the level set.}
    \label{fig:1d_latent}
\end{figure}

When considering multiple sensor positions, the prior mean corresponds to the maximum of the individual position-induced priors:
\begin{equation}
   m(x) = \underset{i}{\max} \{\gamma_i k(s_i, x)\}.
   \label{eq:gp_mean}
\end{equation}
Note that each sensor pose can be associated with a different maximum range $\left(\gamma_i = \frac{c}{\kappa(\range_{\max,i})}\right)$.
Fig.~\ref{fig:1d_latent} shows the field behaviour with 3 poses and 2 point measurements.
As specified in Section~\ref{sec:problem_statement}, the field crosses the level set at the boundary between free and unknown space.
This transition can correspond to the presence of a surface in the environment or the frontier between already observed and not-yet-explored areas (e.g., at the maximum sensor range).
Discriminating between these two situations can be done using a simple threshold on the inferred variance.

\subsection{Behaviour in higher dimensions and bubble prior}

\begin{figure}
    \centering
    \def\vdist{-0.1cm}
    \def\vdistbis{-0.0cm}
    \def\hdist{0.0cm}
    \def\largevdist{0.6cm}
    \def\imgheight{3.9cm}
    \def\toplegenddist{0.5cm}
    \def\legenddist{0.1cm}
    \def\linedist{0.35cm}
    \def\lineleft{2.4cm}
    \def\lineright{1.85cm}
    \begin{tikzpicture}
        \tikzstyle{image} = [inner sep=0, outer sep=0]
        \tikzstyle{legend} = [inner sep=0, outer sep=0, execute at begin node=\setlength{\baselineskip}{8pt} \scriptsize]
        \tikzstyle{toplegend} = [inner sep=0, outer sep=0, execute at begin node=\setlength{\baselineskip}{8pt} \small]

        \node[image] (priorB) {\includegraphics[clip,height=\imgheight, trim=0.9cm 1cm 1.0cm 1.6cm]{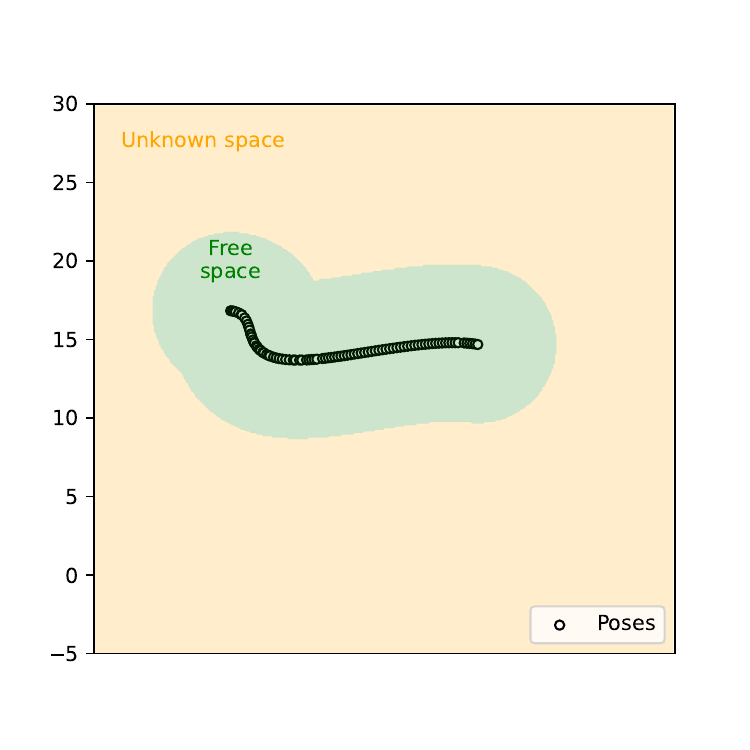}};
        \node[image, below=\vdist of priorB] (priorC) {\includegraphics[clip,height=\imgheight, trim=0.9cm 1cm 1.0cm 1.6cm]{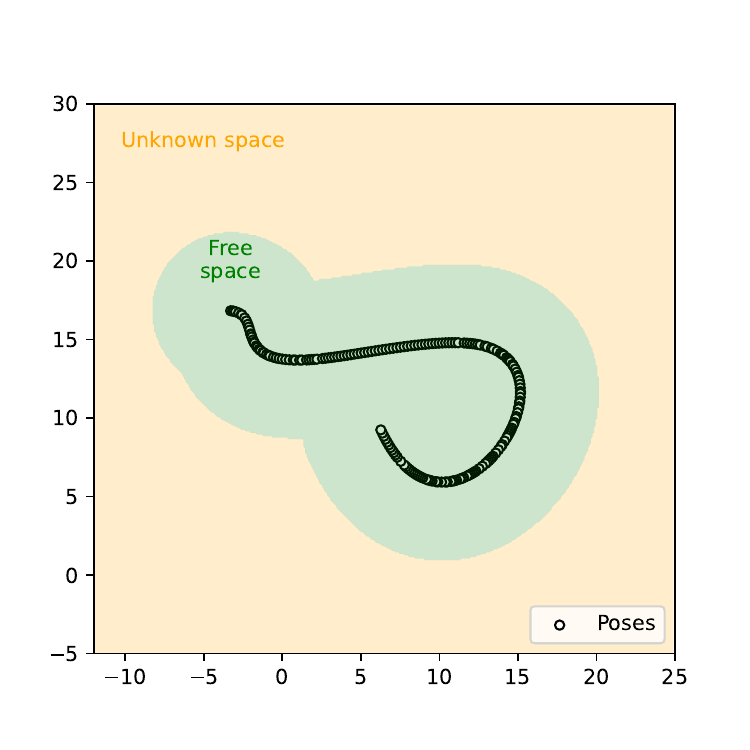}};
        \node[image, right=\hdist of priorB] (postB) {\includegraphics[clip,height=\imgheight, trim=0.9cm 1cm 1.0cm 1.6cm]{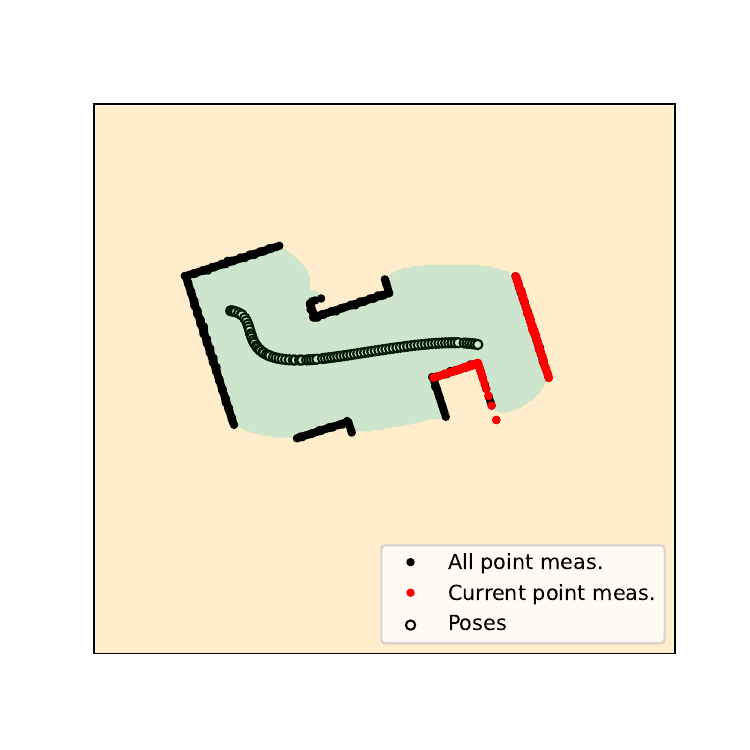}};
        \node[image, below=\vdist of postB] (postC) {\includegraphics[clip,height=\imgheight, trim=0.9cm 1cm 1.0cm 1.6cm]{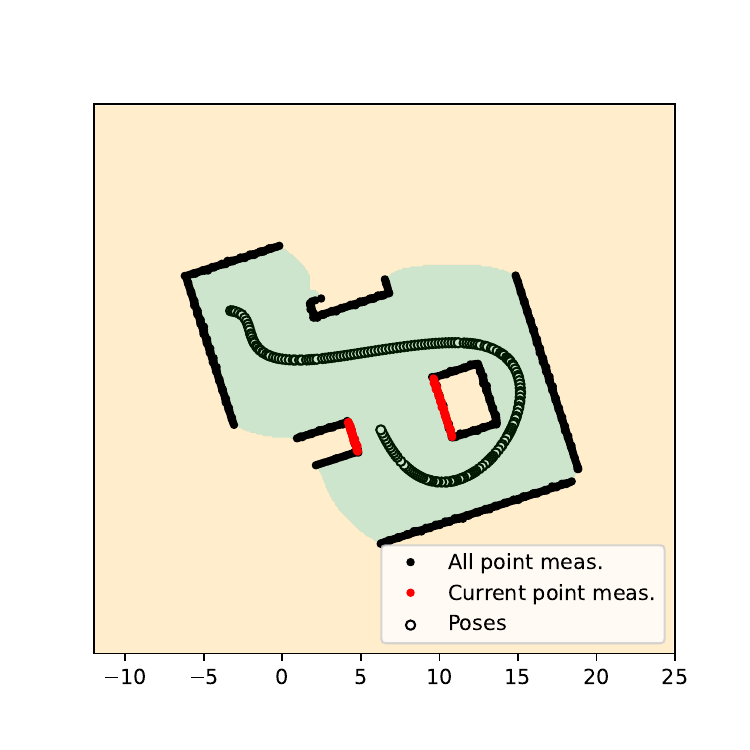}};
        
        \node[image, below=\largevdist of priorC] (priorE) {\includegraphics[clip,height=\imgheight, trim=0.9cm 1cm 1.0cm 1.6cm]{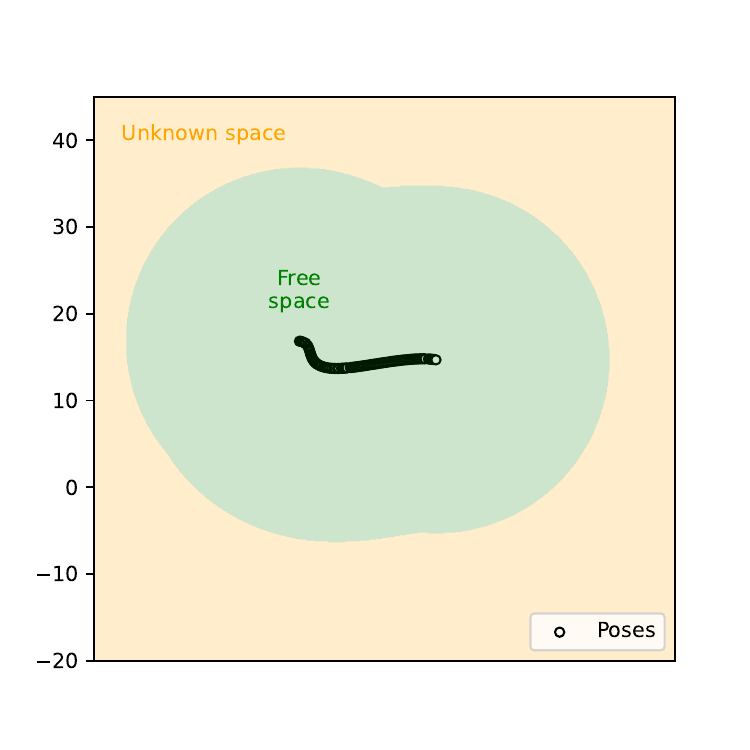}};
        \node[image, below=\vdistbis of priorE] (priorF) {\includegraphics[clip,height=\imgheight, trim=0.9cm 1cm 1.0cm 1.6cm]{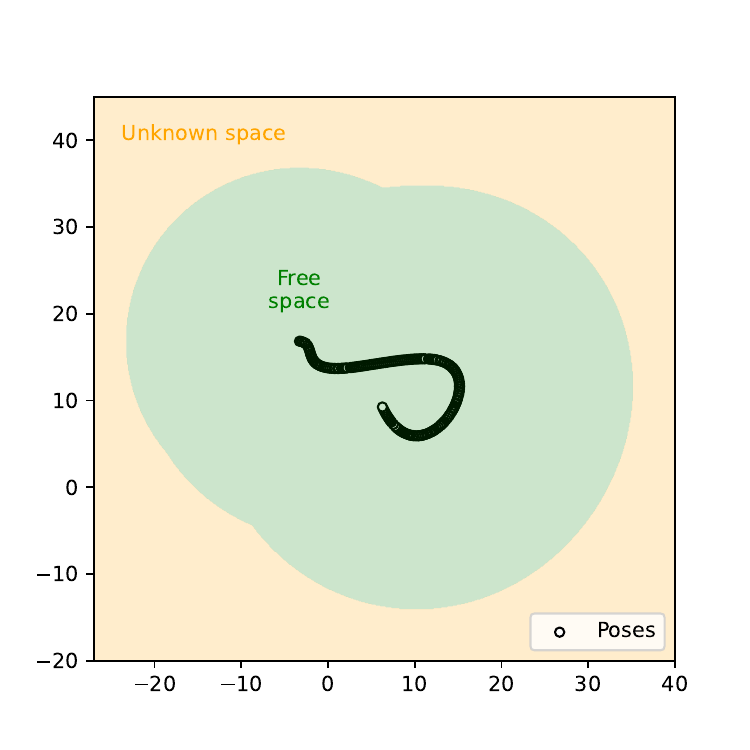}};
        \node[image, right=\hdist of priorE] (postE) {\includegraphics[clip,height=\imgheight, trim=0.9cm 1cm 1.0cm 1.6cm]{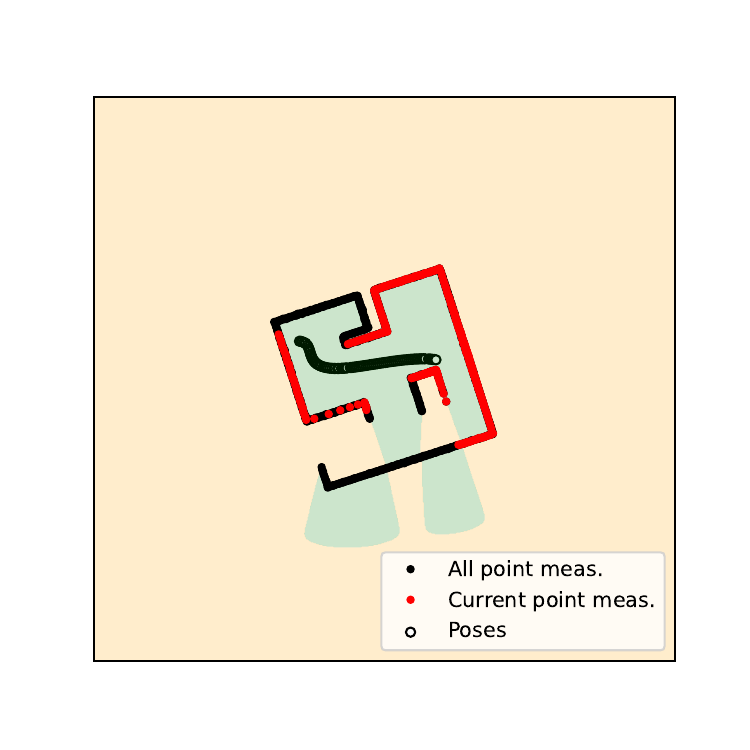}};
        \node[image, below=\vdistbis of postE] (postF) {\includegraphics[clip,height=\imgheight, trim=0.9cm 1.0cm 1.0cm 1.6cm]{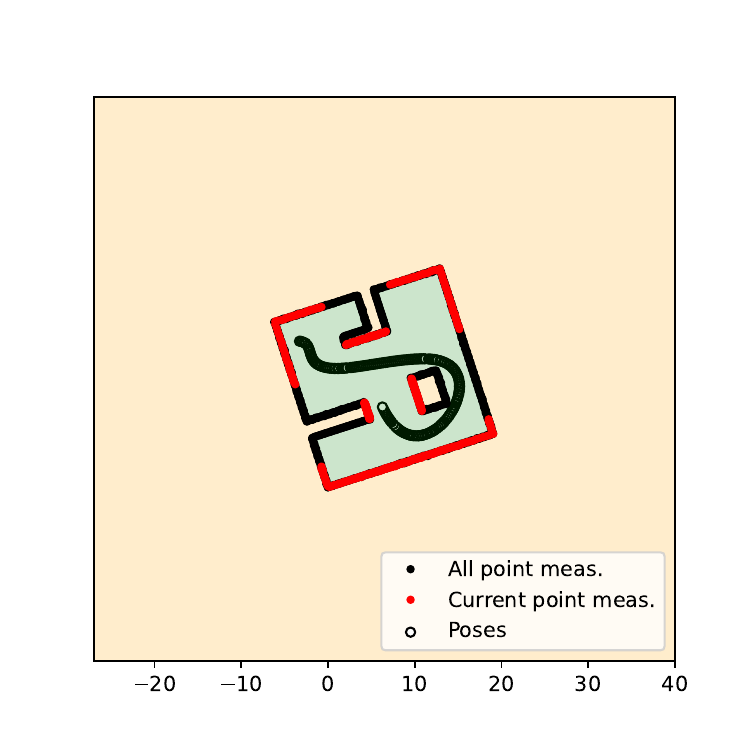}};

        \node[toplegend, above=\toplegenddist of priorB]{\textbf{Prior occupancy}};
        \node[toplegend, above=\toplegenddist of postB]{\textbf{Posterior occupancy}};

        \node[legend, above=0.5*\imgheight + \legenddist of {$(priorB)!0.5!(postB)$}]{$\range_{\max} = 5\,\si{m}$};
        \node[legend, above=0.5*\imgheight + \legenddist of {$(priorE)!0.5!(postE)$}]{$\range_{\max} = 20\,\si{m}$};

        \coordinate[above=\linedist of priorB, xshift=-\lineleft] (lineAleft);
        \coordinate[above=\linedist of postB, xshift=\lineright] (lineAright);
        \draw (lineAleft) -- (lineAright);

        \coordinate[above=\linedist of priorE, xshift=-\lineleft] (lineBleft);
        \coordinate[above=\linedist of postE, xshift=\lineright] (lineBright);
        \draw (lineBleft) -- (lineBright);

        \node[legend, left=\legenddist of priorB, rotate=90, anchor=south] {Time A};
        \node[legend, left=\legenddist of priorC, rotate=90, anchor=south] {Time B};
        \node[legend, left=\legenddist of priorE, rotate=90, anchor=south] {Time A};
        \node[legend, left=\legenddist of priorF, rotate=90, anchor=south] {Time B};
    \end{tikzpicture}
    \caption{Illustration of the pose-based prior and GP-based posterior for occupancy mapping. Our approach handles partial observations appropriately when the prior is close enough to the desired occupancy (upper half). However, when the prior is too far from the target (lower half), the GP fails to infer the occupancy accurately, indicating the need for a more complex prior.}
    \label{fig:2d_fail}
\end{figure}
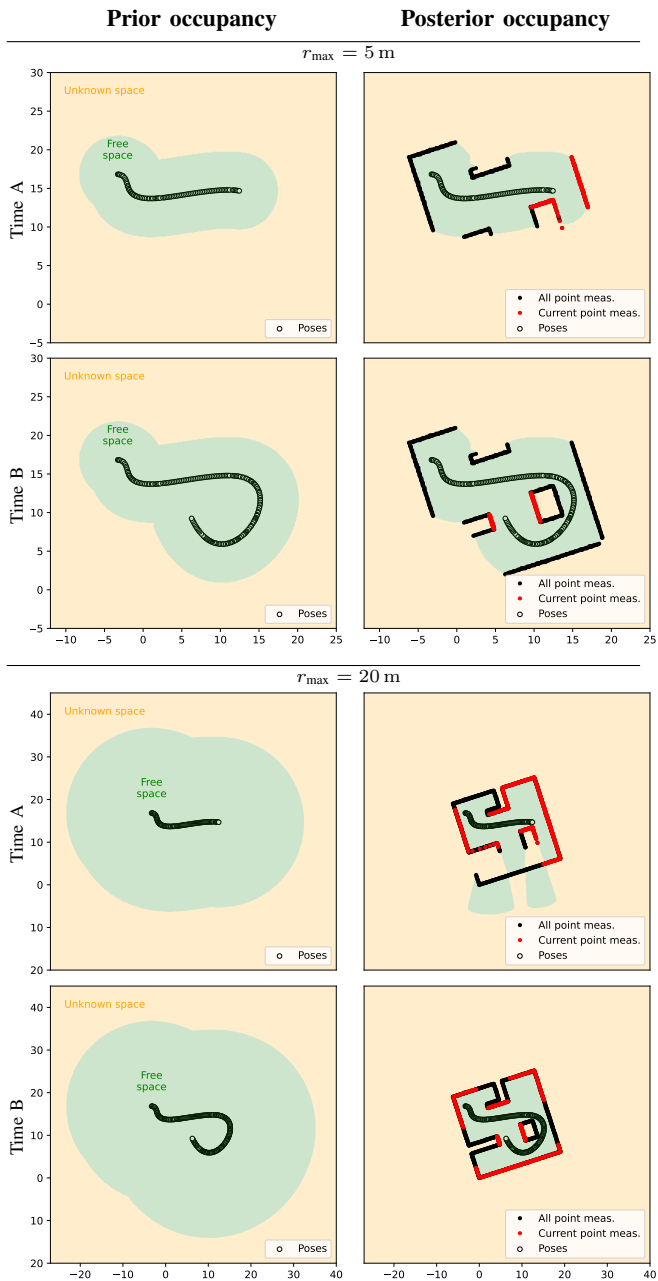

While our derivations leveraged a simple 1D scenario, we show empirically that the desired binarized field behaviour specified in Section~\ref{sec:problem_statement} can be observed in more complex setups, as illustrated in 2D with Fig.~\ref{fig:teaser}.
However, as shown in Fig.~\ref{fig:2d_fail}, our approach can fail if the discrepancy between the prior mean function and the targeted field is too large (only when given partial observations of the environment, third row).

\begin{figure}
    \centering
    \includegraphics[width=0.99\columnwidth]{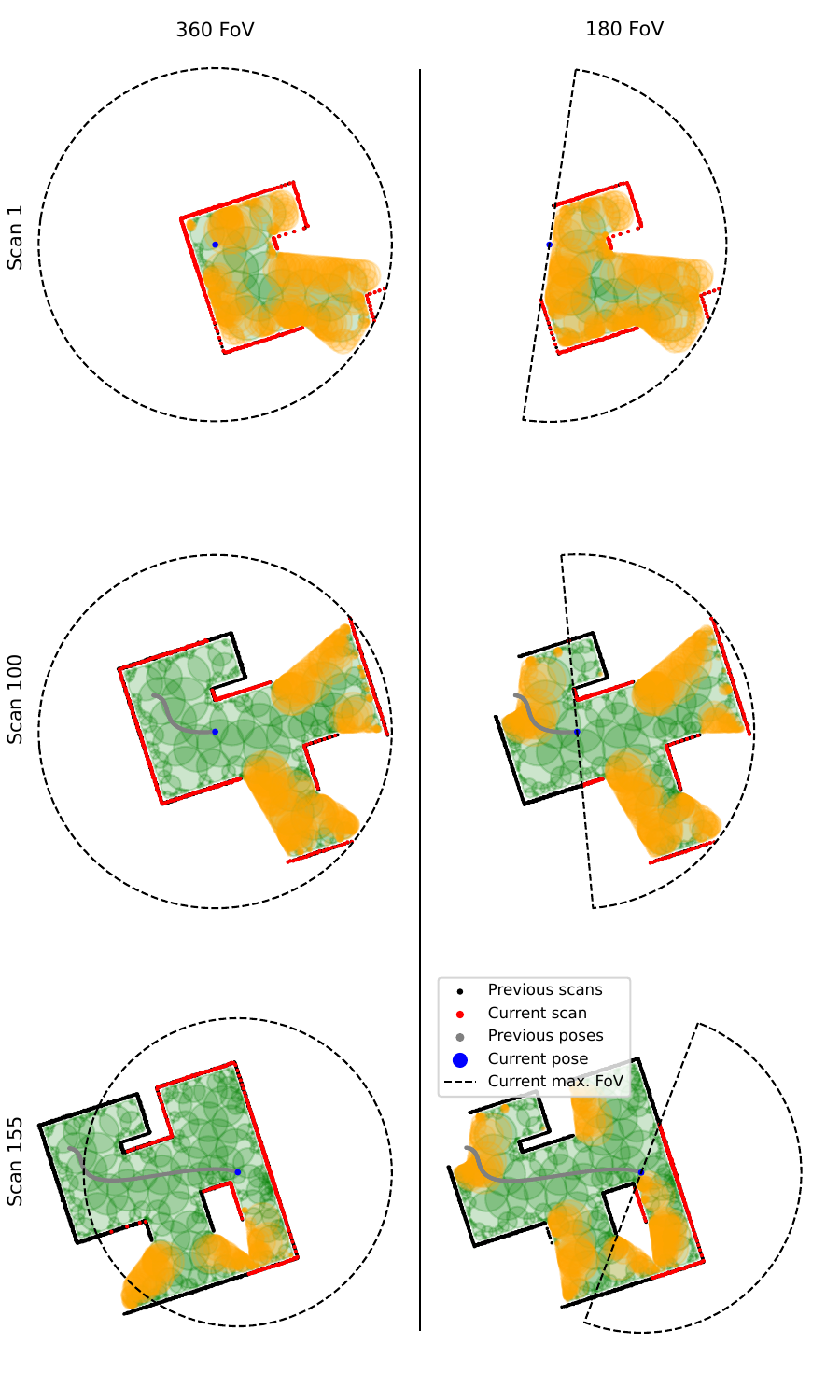}
    \caption{Illustration of the incremental \emph{bubble} coverage used in the proposed bubble-based mean function. The orange bubbles represent bubbles that can be used for coverage expansion (i.e., close to a potential exploration frontier), while the green ones are fixed. Using the bubbles as pseudo poses allows for the computation of a prior mean that is closer to the expected occupancy (compared to the direct use of the sensor maximum range in Fig.~\ref{fig:2d_fail}).}
    \label{fig:bubble_intuition}
\end{figure}

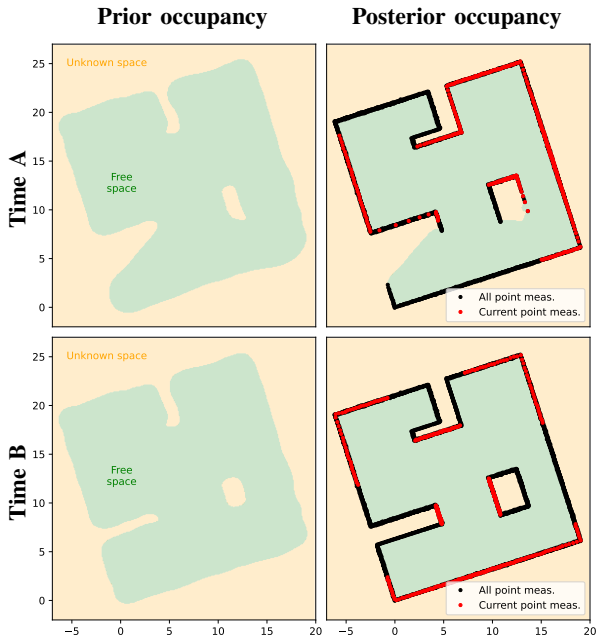
\begin{figure}
    \centering
    \def\vdist{-0.1cm}
    \def\vdistbis{-0.0cm}
    \def\hdist{-0.2cm}
    \def\largevdist{0.6cm}
    \def\imgheight{4.0cm}
    \def\toplegenddist{0.2cm}
    \def\legenddist{0.1cm}
    \def\linedist{0.35cm}
    \def\lineleft{2.4cm}
    \def\lineright{1.85cm}
    \begin{tikzpicture}
        \tikzstyle{image} = [inner sep=0, outer sep=0]
        \tikzstyle{legend} = [inner sep=0, outer sep=0, execute at begin node=\setlength{\baselineskip}{8pt} \scriptsize]
        \tikzstyle{toplegend} = [inner sep=0, outer sep=0, execute at begin node=\setlength{\baselineskip}{8pt} \small]

        \node[image] (postA) {\includegraphics[clip,height=\imgheight, trim=1.3cm 0.8cm 1.4cm 1.5cm]{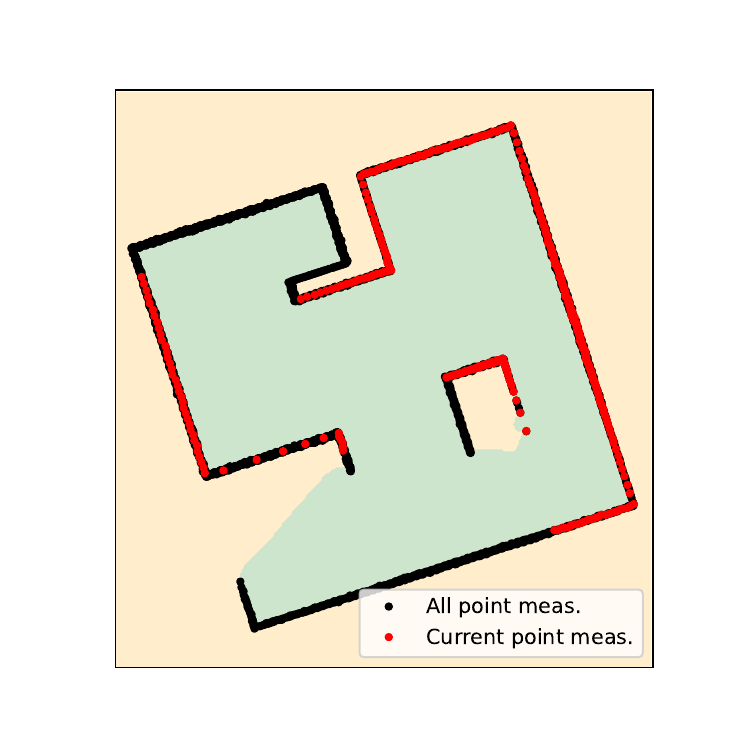}};
        \node[image, below=\vdist of postA] (postB) {\includegraphics[clip,height=\imgheight, trim=1.3cm 0.8cm 1.4cm 1.5cm]{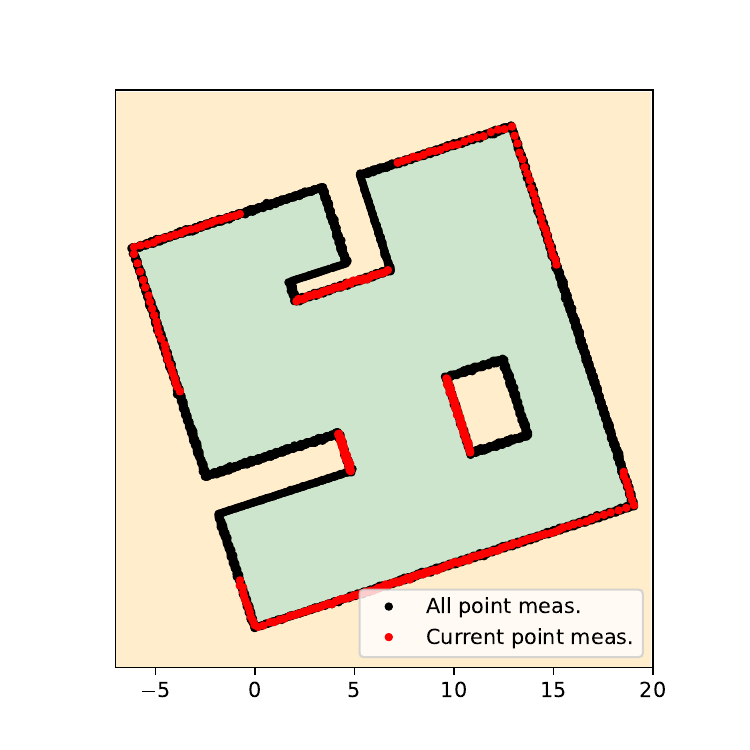}};
        \node[image, left=\hdist of postA] (priorA) {\includegraphics[clip,height=\imgheight, trim=1.3cm 0.8cm 1.4cm 1.5cm]{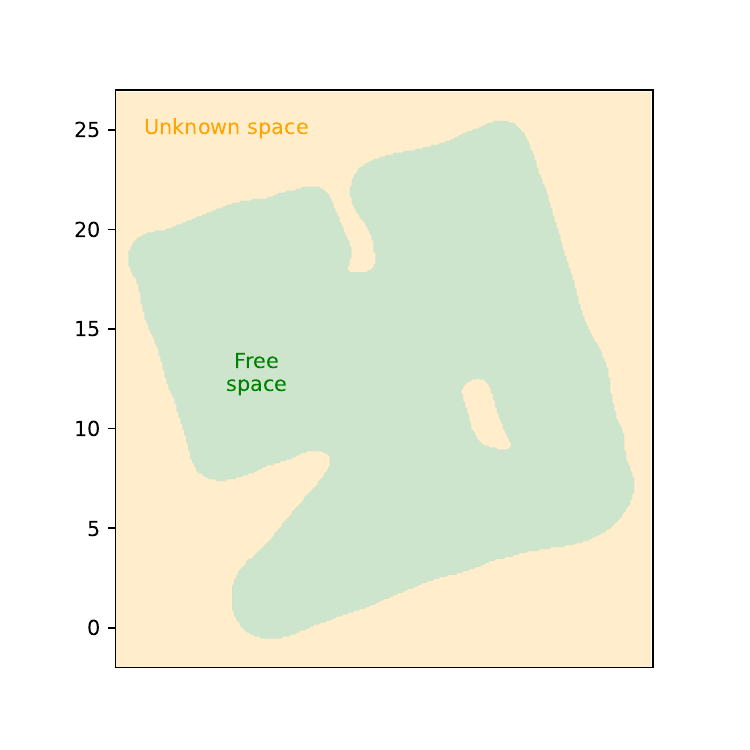}};
        \node[image, below=\vdist of priorA] (priorB) {\includegraphics[clip,height=\imgheight, trim=1.3cm 0.8cm 1.4cm 1.5cm]{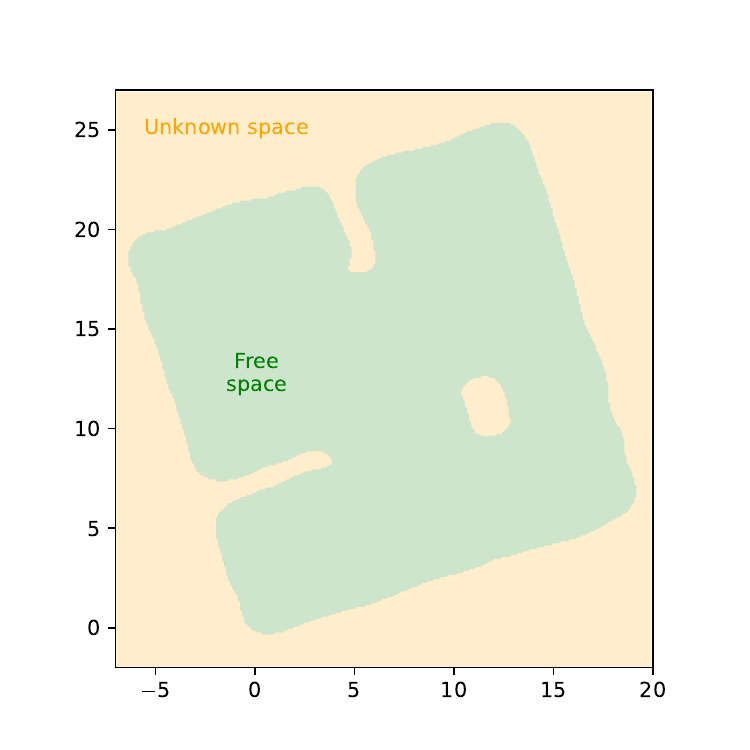}};
        \node[toplegend, above=\toplegenddist of priorA]{\textbf{Prior occupancy}};
        \node[toplegend, above=\toplegenddist of postA]{\textbf{Posterior occupancy}};
        \node[toplegend, left=\legenddist of priorA, rotate=90, anchor=south] {\textbf{Time A}};
        \node[toplegend, left=\legenddist of priorB, rotate=90, anchor=south] {\textbf{Time B}};
    \end{tikzpicture}
    \caption{Leveraging our bubble-based occupancy prior allows the adequate inference of the GP posterior occupancy (right), when using solely the robot's pose led to failure (cf. Fig.~\ref{fig:2d_fail}).}
    \label{fig:bubble}
\end{figure}

To address this issue, we introduce a more complex representation of the sensor's \ac{fov}.
Instead of considering a single position and range limit, we fill the sensor's \ac{fov} with multiple \emph{bubbles} that have their own position and range limit.
Using an incremental algorithm similar to the one introduced in~\cite{lee2024safe} (illustrated in Fig.~\ref{fig:bubble_intuition}) allows us to compute a prior that is closer to the targeted field while accommodating for arbitrarily shaped \acp{fov}.
The bubbles' radii are computed to keep a predefined clearance from the environment's objects using an Euclidean distance field derived from the point measurements.
From the bubble coverage, the prior mean is computed as before in \eqref{eq:gp_mean} but replacing the set of sensor poses and their associated maximum range by the bubble centres and their radii (plus 2 times the clearance).
Fig.~\ref{fig:bubble} shows the resulting prior mean and the final field inference at two different timestamps.
While this representation is denser than the original pose-only derivation, it stays sparse as it is not tied to the kernel's length-scale, the number of bubbles is contained ($<400$ in our simulated environment), and they do not impact the \ac{gp} cubic computational complexity as they are not considered observations.
Furthermore, the growing algorithm can limit the size of the bubbles with an arbitrary maximum radius, thus, the field inference can be performed using solely local information by using simple radius searches.
Therefore, this approach alleviates the overall cubic complexity of \ac{gp} regression, provided an efficient implementation with appropriate data structures (cf. Section~\ref{sec:implementation}).

\section{Experiments}
\label{sec:experiments}

\subsection{Implementation}
\label{sec:implementation}

We have implemented the proposed method in C++.
Our implementation performs incremental occupancy mapping using a stream of point clouds and poses as input.
For the sake of efficiency, point measurements are stored as voxel centroids in a hash map associated with a spatial index.
Both structures need to allow for the fast insertion of new points.
Thus, as in~\cite{legentil20242fast}, we choose Ankerl's hash map~\cite{ankerl2022hasmapgit} and PhTrees~\cite{zaschke2014phtree}.

Our bubble-growing algorithm is an incremental, deterministic version of the \textit{expansive bubble graph} from~\cite{lee2024safe}, using the accurate \ac{gp}-based distance field~\cite{legentil2024accurate}.
We made the distance field very efficient by sharing the aforementioned data structures and performing local \ac{gp} inference.
A second PhTree stores the bubbles for fast evaluation of the prior mean function through log(n) radius searches.

Briefly, the construction of the proposed occupancy field upon arrival of new range data is
\begin{itemize}
    \item The expansion of the bubble coverage (using a distance field associated with the scan at hand),
    \item The insertion of the new data in the hash map and PhTrees.
\end{itemize}
Querying the field is done by
\begin{itemize}
    \item Computing the prior mean value (radius search in the bubble PhTree),
    \item Querying the closest point-measurement voxel and training a local \ac{gp} using the voxel's neighbourhood.
    \item \ac{gp} mean and variance inference \eqref{eq:gp_posterior}.
\end{itemize}
To perform surface reconstruction, we
\begin{itemize}
    \item Execute a tailored version of the marching cube algorithm~\cite{lorensen1987marching},
    \item Keep solely elements whose vertices' variance is under a certain threshold.
\end{itemize}

\subsection{Reconstruction evaluation in simulation}

\begin{table}
    \centering
    \caption{Analysis of the reconstruction error.}
    \setlength{\tabcolsep}{2pt}
    \begin{tabularx}{\linewidth}{lYY}
        \toprule
        \textbf{Method} & \textbf{Environment A} & \textbf{Environment B} \\
        \midrule
        \textbf{Raw data} & 19.9 / 25.6 & 19.4 / 25.1 \\
        \textbf{Occ. grid} & 23.2 / 27.3 & 22.4 / 23.8 \\
        \textbf{BHM} \cite{senanayake2017bayesian} & 350 / 361 & 210 / 218 \\
        \textbf{iGPOM} \cite{ghaffari2018gaussian} & 114 / 125 & 92.3 / 104 \\
        \textbf{Ours} & \textbf{7.37} / \textbf{9.14} & \textbf{6.63} / \textbf{11.0} \\
        \bottomrule
        \multicolumn{3}{c}{\scriptsize Mean point-to-surface distance / Root Mean Square point-to-surface distance [$\si{mm}$].}
        \\
        \multicolumn{3}{p{\columnwidth}}{\scriptsize $^*$ We have manually removed surface elements that correspond to free/occupied transitions in non-observed areas.}
    \end{tabularx}
    \label{tab:2drmse}
\end{table}

We benchmark our method against a discrete occupancy grid, iGPOM~\cite{ghaffari2018gaussian}, and BHM~\cite{senanayake2017bayesian}.
Using the \ac{gp} occupancy map concept from~\cite{ghaffari2014exploration}, iGPOM maps the environment incrementally.
BHM stands for Bayesian Hilbert mapping and is an extension of the continuous Hilbert maps~\cite{ramos2016hilbert} to account for moving objects in the environment.
Note that we use BHM to process the full batch of data in a single step to optimize its accuracy and computation time.
To evaluate each method's ability to represent obstacles in the scene, we benchmark the reconstruction ability of all methods in simulated environments by comparing the estimated location of walls (transition between free and occupied space) with respect to the environment's ground-truth geometry.
For the occupancy grid approach, the `reconstruction' consists simply of all the cells classified as occupied.

\begin{figure*}
    \centering
    \includegraphics[width=0.99\linewidth, clip, trim=0.2cm 0.3cm 0.2cm 0.3cm]{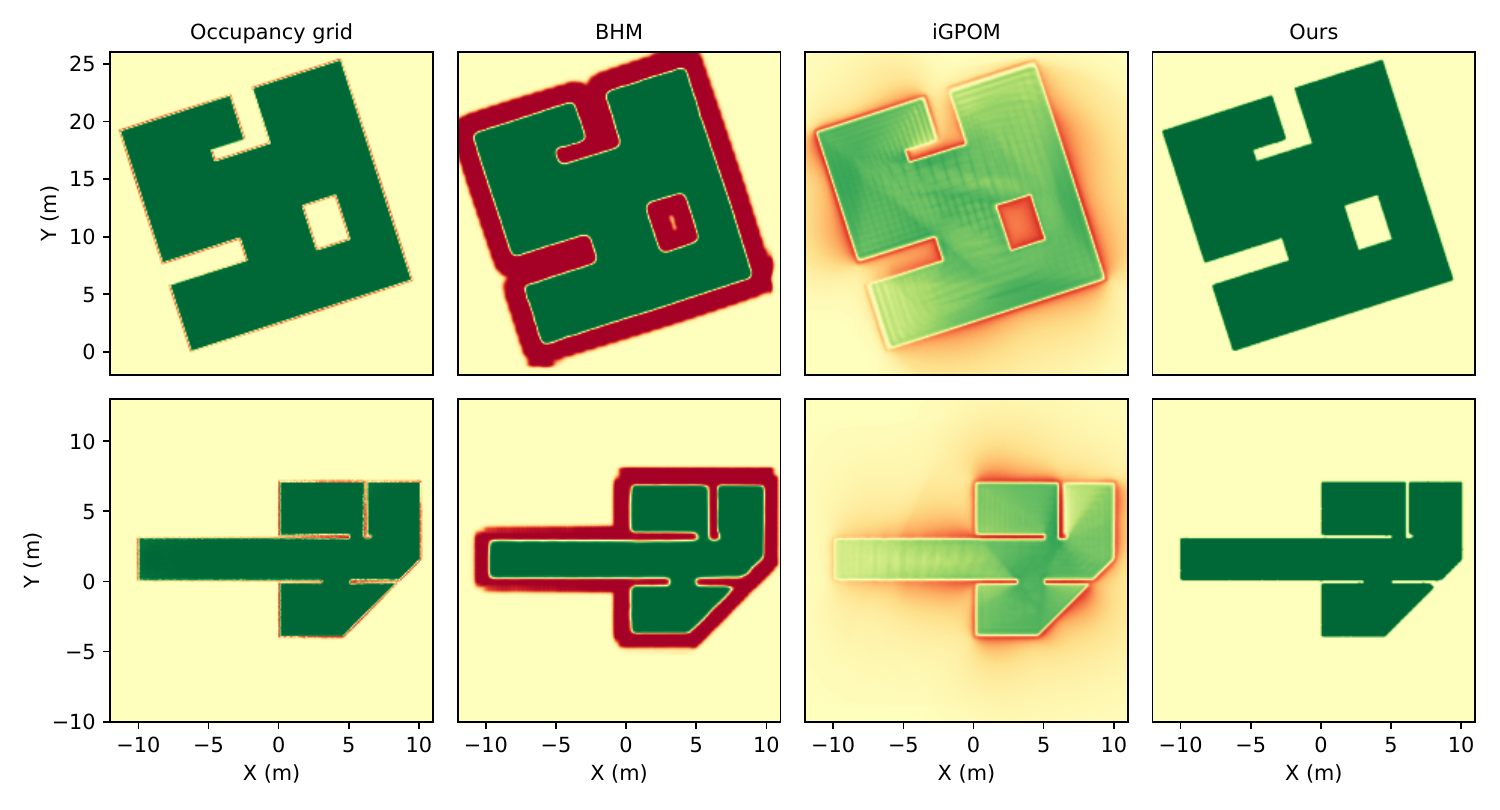}
    \caption{Inferred occupancy using simulated data.}
    \label{fig:2d_simu}
\end{figure*}

Table~\ref{tab:2drmse} shows the RMSE distance-to-surface obtained with all the methods.
Except for ours, all the methods show mean and \ac{rmse} errors superior to the raw point cloud.
This can easily be explained by the fact that these methods perceive obstacles as areas instead of infinitesimally thin surfaces, as illustrated in Fig.~\ref{fig:2d_simu}.
In all fairness, the occupancy methods benchmarked do not specifically address the reconstruction task.
Future work will include additional evaluations with reconstruction-focused methods.
Nonetheless, this metric shows how the environment representations of the benchmarked methods are biased.
Notably, large non-observed areas are classified as occupied by iGPOM and BHM.

Among the continuous methods, ours is by far the most computationally efficient.
The incremental bubble-growing algorithm and the data structure update took on average $2.5\,\si{ms}$ per lidar scan using a single core of an Intel i7-13850HX CPU.
The final reconstruction took less than $0.1\,\si{s}$.
Leveraging multi-threading and \ac{gp} computation caching, the field inference with its variance takes $0.5\,\si{\micro\s}$ per query point on average.
For comparison, the publicly available implementations of BHM and iGPOM are up to four orders of magnitude slower overall: $23.1\,\si{\s}$ per scan for iGPOM on a single core, and $151\,\si{ms}$ (dividing the full batch processing time by the number of scans) for BHM using all CPU cores.

\subsection{Real data}

\begin{figure*}
    \centering
    \includegraphics[width=0.99\linewidth,clip, trim=0.3cm 0.1cm 0.3cm 0.5cm]{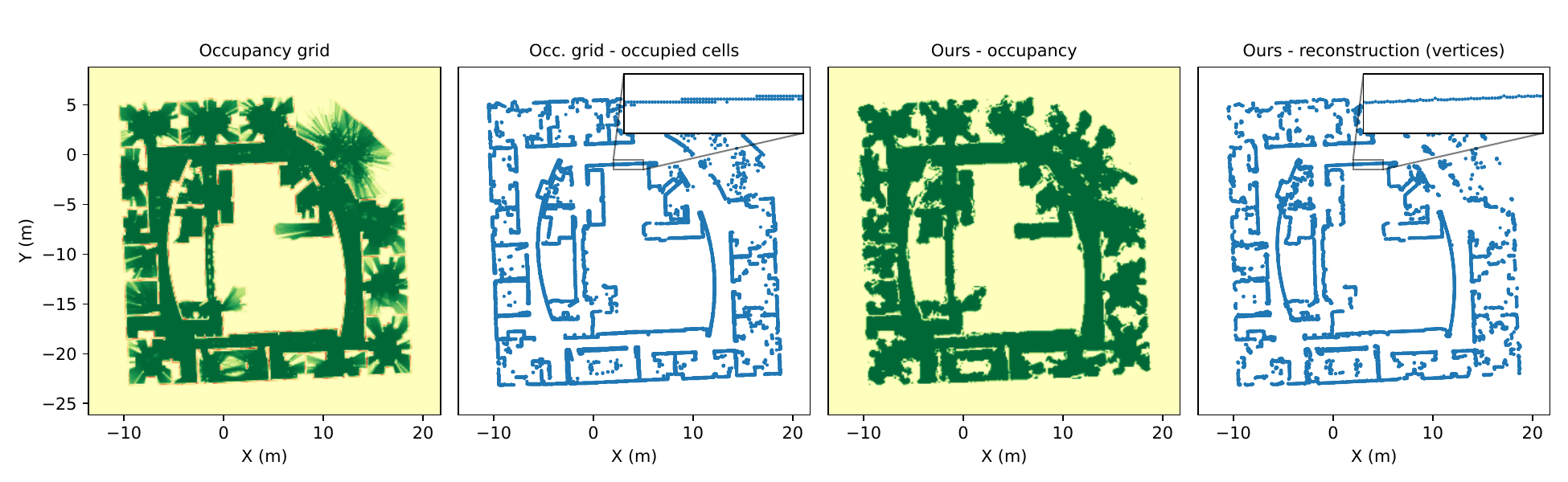}
    \caption{Inferred occupancy and reconstruction using real-world data from the Intel Lab SLAM dataset.}
    \label{fig:2dreal}
\end{figure*}

To the best of our knowledge, no public real-world 2D lidar dataset provides the ground-truth of both the sensor position and the surface geometry.
Nonetheless, we display in Fig.~\ref{fig:2dreal} qualitative results obtained on the Intel Lab 2D SLAM dataset.
Similarly to the simulated scenario, our method is quite efficient and provides an adequate occupancy representation of the environment.
Our reconstruction includes the low-uncertainty structural elements of the environment.
As shown in the zoom insets, the occupied cells of the occupancy grid span over areas with non-zero thickness, while our method infers walls as thin surfaces.

\section{Limitations}

We have demonstrated the soundness of the proposed method with various experiments, but our approach possesses some limitations.
Firstly, compared to the other continuous methods in our benchmark, ours relies on more parameters owing to the bubble-growing algorithm.
These parameters cannot be elegantly and automatically optimized from sample data as in standard \ac{gp} regression.
However, they have straightforward physical meaning (minimum radius, wall clearance, ...) and can easily be set manually without a deep understanding of the method.
The second limitation is the non-smooth nature of the \matern $\nu=0.5$ kernel.
Accordingly, handling highly noisy data might limit the reconstruction accuracy of our method.
Further analysis is required to characterize the method's performance with varying noise levels.
Tangentially, the non-smoothness of the kernel might limit the ability to accurately account for input noise (i.e., using uncertain robot poses) through kernel linearization~\cite {mchutchon2011gaussian}.

\section{Conclusion}

This paper introduces a novel approach to occupancy mapping using \ac{gp} regression.
Unlike previous \ac{gp}-based formulations, our method does not explicitly require free-space observations as we embed the information about the sensor's \ac{fov} within the prior mean function of the \ac{gp} based on the sensor pose and maximum range.
Our approach's theory has been derived using a simple 1D scenario.
We empirically show that the desired behaviour of the field can be observed at higher dimensions in some specific situations where the prior mean does not differ too much from the targeted field.
To address the non-compliant scenarios and to accommodate for non-omnidirectional \ac{fov}, we leverage a bubble-growing algorithm to model the sensor's \ac{fov} efficiently.
Given the use of the right data structures, this latter addition also allows highly efficient field inference as the required information is present locally, thus only requiring local \ac{gp} training.

Our future work aims toward efficient and accurate 3D surface reconstruction.
Thus, we will implement an efficient 3D marching cube algorithm tailored to the proposed field and data structure.
It will allow reconstruction comparison with grid-occupancy, \ac{sdf}, and learning-based approaches.

\bibliographystyle{plainnat}
\bibliography{bibliography}

\newpage
\appendices
\section{\matern 0.5 kernel properties}
\label{app:matern}

This appendix shows how using the \matern $\nu=0.5$ kernel with our \ac{gp}-based field formulation complies with properties (i) and (ii) from Section~\ref{sec:derivation} when dealing with a 1D environment.
Considering a single pose $s$ and a maximum range $\range_{\max}$, property (i) corresponds to
\begin{equation}
    m(x) \begin{cases}
        =c & \text{for}\ x=s-\range_{\max}\ \text{or}\ x=s+\range_{\max} 
        \\
        > c & \text{for}\ s-\range_{\max} < x < s+\range_{\max}
        \\
        < c & \text{otherwise}
    \end{cases}
    .
    \nonumber
\end{equation}
As illustrated in Fig.~\ref{fig:matern} (left), choosing 
\begin{equation}
    m(x)=\gamma k(s,x)\ \ \text{with}\ \ \gamma = \frac{c}{\kappa(\range_{\max})}
\label{eq:meanfun}
\end{equation}
fulfills property (i) when using the unscaled \matern 0.5 kernel $k(x, x') = \kappa(\vert x - x'\vert) = \exp(-\frac{\vert x - x'\vert}{l}))$  where $l$ is the lengthscale parameter.

When considering a pose $s$ and a point $p$ (due to the sensor's maximum range, $s-\range_{\max} < p < s+\range_{\max}$), property (ii) means that the latent field on the unobserved side of $p$ ($x \geq p$ if $p > s$, $x \leq p$ otherwise) should be a translation of the mean function for $x \geq s + \range_{\max}$ (or $x \leq s - \range_{\max}$ when $p < s$). In Fig.~\ref{fig:matern}, it means that the latent field for $x > 3$ on the left and for $x>2.5$ on the right should be equal.
Formally, property (ii) is fulfilled if $m(s+\range_{\max} + x) = f(p+x)$ for $x>0$ and $p > s$.
The left-hand side can be rewritten with~\eqref{eq:meanfun} as
\begin{equation}
m(s+\range_{\max} + x) = \gamma \kappa(\range_{\max} + x).
\nonumber
\end{equation}
For the right-hand side, let us use~\eqref{eq:gp_posterior} with a noiseless measurement:
\begin{align}
    \begin{aligned}
    f(p+x)&= m(p+x) + k(p, p+x)(c-m(p))\\
    & = \gamma\kappa(p+x-s) + \kappa(x)(\gamma\kappa(\range_{\max}) - \gamma\kappa(p-s)).
    \end{aligned}
    \nonumber
\end{align}
By developing the product and leveraging the fact that using the \matern 0.5 kernel $\kappa(x)\kappa(p-s) = \kappa(x+p-s)$ and $\kappa(x)\kappa(\range_{\max}) = \kappa(x+\range_{\max})$ (as $x>0$, $p-s >0$, and $\range_{\max}$), we obtain
\begin{equation}
    f(p+x)=\gamma \kappa(\range_{\max} + x).
    \nonumber
\end{equation}

\end{document}